\pdfoutput=1

\documentclass[11pt]{article}

\usepackage{ACL2023}

\usepackage{times}
\usepackage{latexsym}
\usepackage{tabularx,booktabs}
\usepackage{color,soul}
\usepackage{multirow}
\usepackage{graphicx}
\usepackage{makecell}

\usepackage[T1]{fontenc}

\usepackage[utf8]{inputenc}

\usepackage{microtype}

\usepackage{inconsolata}

 \makeatletter
\def\@fnsymbol#1{\ensuremath{\ifcase#1\or \dagger\or \ddagger\or
   \mathsection\or \mathparagraph\or \|\or **\or \dagger\dagger
   \or \ddagger\ddagger \else\@ctrerr\fi}}
    \makeatother
    
\newcommand{\autoevals}{evaluation metrics}

\title{MISMATCH: Fine-grained Evaluation of Machine-generated Text \\ with Mismatch Error Types}

\author{ Keerthiram Murugesan \thanks{~~IBM Research. Correspondence to: Keerthiram Murugesan $<$keerthiram.murugesan@ibm.com$>$. Code available at \url{https://github.com/IBM/mismatch-eval}} \And Sarathkrishna Swaminathan\footnotemark[1] \And Soham Dan\footnotemark[1] \AND
Subhajit Chaudhury\footnotemark[1] \And Chulaka Gunasekara\footnotemark[1] \And Maxwell Crouse\footnotemark[1] \AND
Diwakar Mahajan\footnotemark[1] \And Ibrahim Abdelaziz\footnotemark[1] \And Achille Fokoue\footnotemark[1] \AND
Pavan Kapanipathi\footnotemark[1] \And Salim Roukos\footnotemark[1] \And Alexander Gray\footnotemark[1]}

\begin{document}
\maketitle
\begin{abstract}

With the growing interest in large language models, the need for evaluating the quality of machine text compared to reference (typically human-generated) text has become focal attention. Most recent works focus either on task-specific evaluation metrics or study the properties of machine-generated text captured by the existing metrics. 
In this work, we propose a new evaluation scheme to model human judgments in 7 NLP tasks, based on the fine-grained mismatches between a pair of texts. 
Inspired by the recent efforts in several NLP tasks for fine-grained evaluation, we introduce a set of 13 \textit{mismatch error types} such as spatial/geographic errors, entity errors, etc, to guide the model for better prediction of human judgments. We propose a neural framework for evaluating machine texts that uses these mismatch error types as auxiliary tasks and re-purposes the existing single-number evaluation metrics as additional scalar features, in addition to textual features extracted from the machine and reference texts. 
Our experiments reveal key insights about the existing metrics via the mismatch errors. We show that the mismatch errors between the sentence pairs on the held-out datasets from 7 NLP tasks align well with the human evaluation.

\end{abstract}

\section{Introduction}
\label{sec:introduction}

Large language models have pushed the boundaries for natural language generation (NLG). More and more, the generated machine texts look human-like. The need for evaluation metrics has never been so critical in the recent decade. Typically, there are two ways to evaluate the quality of machine-generated text:  automatic evaluation and human evaluation.
In automatic evaluation, the quality of the machine-generated text is captured using a single number from a range of values indicating how good the generated text is by a (hand-coded rule-based or neural-based) model. Several NLP tasks still use the metrics from 2 decades ago, for instance, \cite{lin2004rouge} and METEOR \cite{banerjee2005meteor} for abstractive summarization, BLEU \cite{papineni2002bleu} for machine translation, etc.

\begin{figure}[t]
    \centering
    \includegraphics[width=0.95\linewidth]{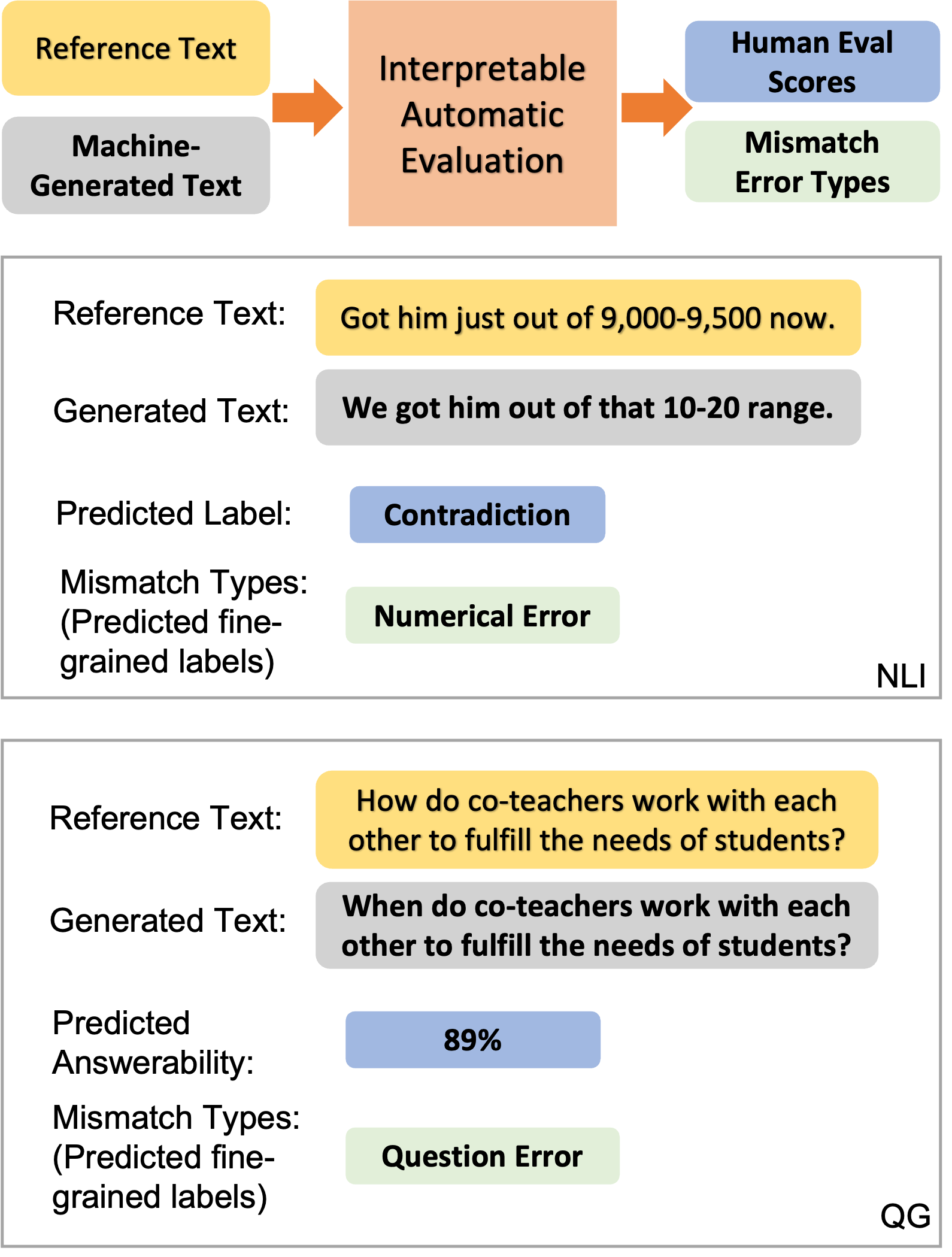}
    \caption{Overview of the proposed fine-grained automatic evaluation of machine-generated text with mismatch error types, along with human evaluation scores. Sample examples are taken from Natural Language Inference (NLI) and Question Generation (QG) tasks.}
    \label{fig:overview}
\end{figure}

It has been noted in several works that automatic evaluation metrics are incapable of capturing the different criteria in measuring the quality of the text and often have a poor correlation with human judgments \cite{sai2021perturbation, callison2006re}.  The current automatic evaluation metrics lack the ability to measure the quality of a modern machine-generated text. In human evaluation, we evaluate the machine text based on human ratings, where we ask human annotators to judge a given pair of texts. The quality of the machine text is measured using different task-specific human evaluation criteria such as fluency, coherence, correctness, consistency, relevance, adequacy, etc. Human evaluations are often expensive, time-consuming, and subjective (low inter-annotator agreement), especially when broad criteria such as the fluency of the generated text and the interestingness of the model-generated text are used for human judgment.

To address these challenges in automatic and human evaluations, there have been recent efforts in the fine-grained evaluation of generated text in several NLP domains \cite{callison2006re,ethayarajh2020utility,sai2021perturbation,see2019makes}. In this paper, we are interested in utilizing fine-grained evaluation categories to guide the prediction of human judgments. Towards this goal, we introduce a task-agnostic list of 13 \textit{mismatch error types}, such as grammatical errors, spatial/temporal errors, etc, that unifies several related task-specific efforts ~\cite{pagnoni2021understanding,glockner2018breaking,dou2022gpt}. These mismatch error types are comprehensive, interpretable, and useful for predicting human evaluation criteria. For example, an occurrence of grammatical error in a machine-generated text can impact its fluency rating.

Figure \ref{fig:overview} gives the overview of the proposed mismatch error types for fine-grained evaluation. We propose a neural framework for evaluation that uses these mismatch error types as auxiliary tasks to model the human judgment and repurposes automated evaluation metrics as additional scalar features, concatenated to textual features extracted from the machine and reference texts via pre-trained LM text embeddings \cite{devlin2019bert}.
We show that pre-training our proposed model using synthetic data for the mismatch prediction task, and fine-tuning using real data for human evaluation criteria, for different NLP tasks, achieves state-of-the-art performance on the main downstream task of predicting human evaluation metrics.
We provide several ablation studies showing the importance of each component of our architecture, and the correlations between the mismatch error types and the automatic and human evaluation metrics. We also show how our architecture is useful in predicting novel evaluation criteria, such as factuality in abstractive summarization.

\section{NLG Evaluation }

\begin{table*}[ht]
\resizebox{\textwidth}{!}{%
\begin{tabular}{llll}
\hline
\textbf{Error Type} & Abbr &
  \textbf{Definition} &
  \textbf{Example Sentence} \\ \hline
\textit{Grammatical/Usage Error} & GramErr &
  Faulty or incorrect use of the grammar and syntax. &
  \begin{tabular}[c]{@{}l@{}}\textbf{ref}:  Two paintings are on the wall.\\ \textbf{gen}: Two painting is on the wall.\end{tabular} \\ \hline
\textit{Predicate Error} & PredErr &
  \begin{tabular}[c]{@{}l@{}}Error in the predicate or its usage \\ with respect to the reference text.\end{tabular} &
  \begin{tabular}[c]{@{}l@{}}\textbf{ref}:  John entered the kitchen.\\ \textbf{gen}: John found the kitchen.\end{tabular} \\ \hline
\textit{Entity Error} & EntErr &
  Mismatch in the primary arguments of the predicate. &
  \begin{tabular}[c]{@{}l@{}}\textbf{ref}:  A dog chased a cat.\\ \textbf{gen}: A dog chased a rat.\end{tabular} \\ \hline
\textit{Predicate Ordering Error} & PredOrdErr &
  \begin{tabular}[c]{@{}l@{}}Error in causal or temporal ordering of \\ the predicates/events.\end{tabular} & 
  \begin{tabular}[c]{@{}l@{}}\textbf{ref}: The police arrested the suspect \\ then he was taken to prison.\\ \textbf{gen}: The suspect was taken to the prison\\ then the police arrested him.\end{tabular} \\ \hline
\textit{Hyponyms/Hypernyms Errors} & HypErr &
  Violations in hypernym/hyponym usage. &
  \begin{tabular}[c]{@{}l@{}}\textbf{ref}: Jim studied mechanical engineering.\\ \textbf{gen}: Jim studied architectural science.\end{tabular} \\ \hline
\textit{Numerical Error} & NumErr &
  \begin{tabular}[c]{@{}l@{}}Error in numerical, quantifiers or related to numbers \\ (ordinals, cardinals, etc)\end{tabular} &
  \begin{tabular}[c]{@{}l@{}}\textbf{ref}:  Martha ate four apples.\\ \textbf{gen}: Martha ate six apples.\end{tabular} \\ \hline
\textit{Spatial/Temporal Error} & STErr &
  \begin{tabular}[c]{@{}l@{}}Error in spatial or geographic information \\ (location, time, etc).\end{tabular} &
  \begin{tabular}[c]{@{}l@{}}\textbf{ref}:  Dave lives in south Chicago.\\ \textbf{gen}: Dave lives in south Chile.\end{tabular} \\ \hline
\textit{Attribute/Modifier Error} & AttrErr &
  \begin{tabular}[c]{@{}l@{}}Mistakes in additional information \\ concerning the predicates and entities.\\ (not covered by numerical, spatial, geographic)\end{tabular} &
  \begin{tabular}[c]{@{}l@{}}\textbf{ref}:  Greg has two small dogs.\\ \textbf{gen}: Greg has two big dogs.\end{tabular} \\ \hline
\textit{Question Error} & QuestErr &
Error/change in the nature of the question's intention. &\begin{tabular}[c]{@{}l@{}}\textbf{ref}:  Did you take the dog to the vet?\\ \textbf{gen}: When did you take the dog to the vet?\end{tabular} \\ \hline
\textit{Negation} & NegErr &
  Negated compared to the reference text. &
  \begin{tabular}[c]{@{}l@{}}\textbf{ref}: Susan took the gift. \\ \textbf{gen}: Susan did not take the gift.\end{tabular} \\ \hline
\textit{Missing Information} & MissInfo &
  Missing key details from the reference text. &
  \begin{tabular}[c]{@{}l@{}}\textbf{ref}: Bob drove to the hospital and saw a doctor.\\ \textbf{gen}: Bob saw a doctor.\end{tabular} \\ \hline
\textit{Out of Reference} & OutofRef &
  Contains additional details not present in the reference text. &
  \begin{tabular}[c]{@{}l@{}}\textbf{ref}: Jack and Jane are friends.\\ \textbf{gen}: Jack and Jane are friends. Jack plays football.\end{tabular} \\ \hline
\textit{Redundant/Repetition} & RepErr &
  Same/similar information repeated more than once. & 
  \begin{tabular}[c]{@{}l@{}}\textbf{ref}:  Tom met Sam at the party.\\ \textbf{gen}: Tom went to the party. Tom met Sam at the party.\end{tabular} \\ \hline
\end{tabular}
}
\caption{Fine-grained evaluation with Mismatch error types between the reference and model-generated texts.}
\label{tab:errortypes_short}
\end{table*}

Given a pair of texts: a reference text and a machine-generated one, we are interested in evaluating the quality of the generated text using the reference text. We measure the quality of the generated text by estimating how a human will judge this text based on different evaluation criteria. Such evaluation is common in many ML/NLP tasks, e.g.,  machine translation, summarization, image captioning, etc. Unlike in other automatic evaluation metrics, we consider fine-grained evaluation cues from $13$ mismatch error types, inspired by several related task-specific efforts ~\cite{pagnoni2021understanding,glockner2018breaking,dou2022gpt} to guide the main task of predicting the human judgments. 
We propose a neural framework for evaluating machine-generated texts that use these mismatch error type predictions as auxiliary tasks, and automated evaluation metrics as additional scalar features, along with the pair of pre-trained LM text embeddings extracted from reference and generated texts. In this section, we discuss the role of mismatch error types as a good proxy for human judgments (Section \ref{sec:mismatch}) and the model architecture for the proposed approach (Section \ref{sec:approach}).

\subsection{Mismatch Error Types}
\label{sec:mismatch}
Recently there has been a growing interest in a set of measurable fine-grained evaluation criteria \cite{dou2022gpt, pagnoni2021understanding, glockner2018breaking}. Most of the recent works require human annotation. In this paper, we consider \textsc{mismatch} types which identify a specific violation or mismatch between a pair of texts spanning various dimensions of semantic structure: whether the mismatch is within a semantic frame, including predicates, entities, modifiers or across multiple semantic frames, for instance predicate ordering mismatch. These mismatch error types can be used as a proxy to measure the broad evaluation categories: a mismatch in sentence ordering can be a weak signal for the coherence of the generated text, and a change in the object names, gender, and numbers can indicate the correctness of the generated text. {Table 1} shows the list of mismatch error types used in this paper. 
We want to understand the relationships between the mismatch types, the \autoevals\ and the human evaluation criteria by addressing the following three questions:
\vspace{-3mm}
\begin{itemize}
 \item  \textit{Are mismatch types a good proxy for human evaluation criteria?} 
We show that the fine-grained evaluation based on the mismatch types can be used to approximate the evaluation criteria used for human ratings.

 \item  \textit{Can we predict a mismatch type between a given pair of texts?}
We demonstrate that, in addition to the BERT-based text representations, \autoevals\ computed from the input pair of texts can  \textit{reliably} identify these mismatch types (with relatively fewer examples for training).

 \item  \textit{Can we use these \autoevals~to predict mismatches on an unseen text pair?} 
We study the predictive power of these \autoevals and demonstrate that even though the \autoevals~do not agree with human evaluation criteria, they can easily identify these mismatch types between pairs of text.
   
\end{itemize} 

\begin{figure*}[ht]
    \centering
    \includegraphics[width=0.75\linewidth]{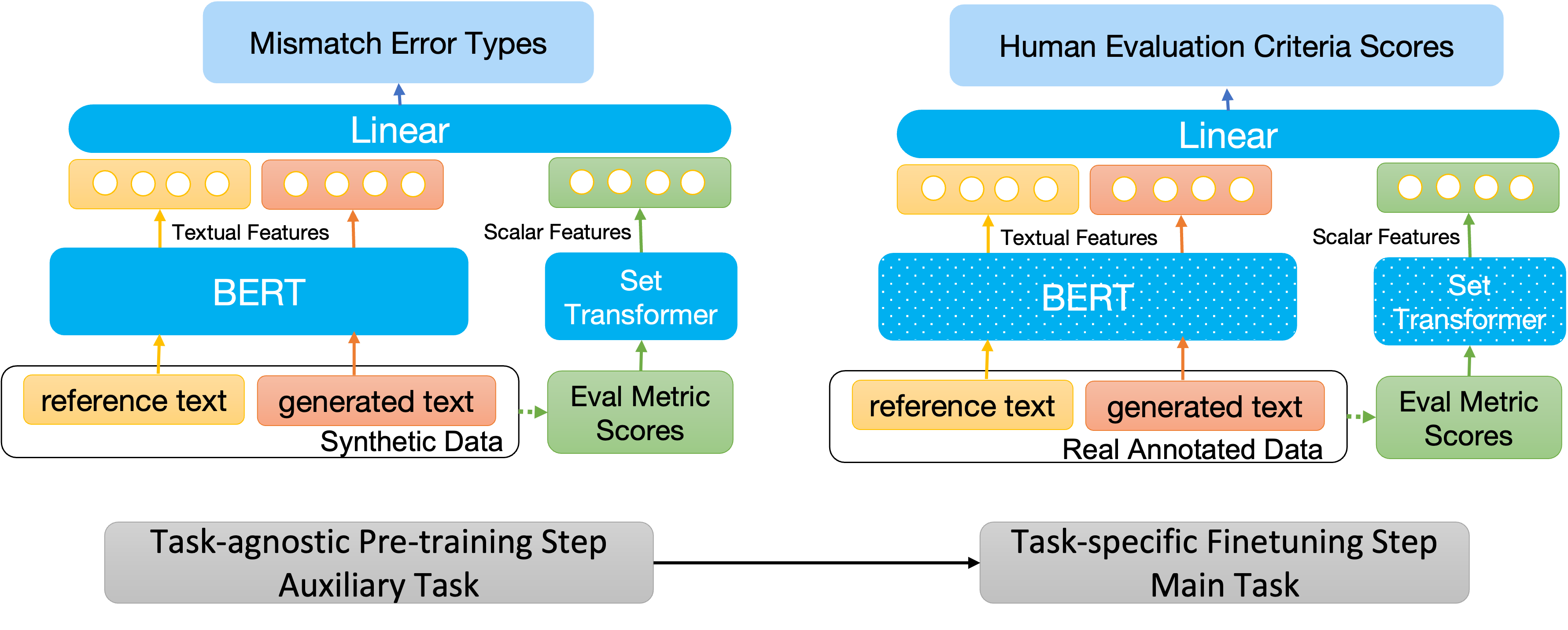}
    \caption{An overview of the mismatch-based evaluation architecture showing task-agnostic pre-training with mismatch error type prediction on synthetic data as the auxiliary task (left) and task-specific finetuning with human evaluation criteria on real annotated data as the main task (right). Dotted arrows indicate that the evaluation metric scores are pre-computed. Dotted blocks indicate that the modules are reused from the pre-training step.}
    \label{fig:arch}
\end{figure*}

As we later see in Figure \ref{fig:corr_autoeval}, our proposed mismatched error types correlate well with both the automatic evaluation metrics as well as human evaluation criteria, demonstrating the relevance of these error types.
In the next section, we show how we model the human ratings on 7 NLP tasks: Abstractive Summarization (AS), Image Caption generation (IC), Question Generation (QG), Machine Translation (MT), Dialogue Generation (DG), Data-to-Text generation (D2T) and Natural Language Inference (NLI) using the mismatch types.

\subsection{Mismatch Error Types for NLG Evaluation}
\label{sec:approach}

We now discuss the neural architecture for the proposed NLG evaluation and show how we model the human judgments using the mismatch error types. 

A simple solution to model the human judgments is to directly train a neural network to learn a function that maps the input pairs of texts to human ratings on the different evaluation criteria, but the amount of human-annotated samples available for training in many NLP tasks is very limited. In this paper, we consider fine-grained evaluation cues based on mismatch error types to guide the model for predicting human judgments.
One of the key advantages of using mismatch error types to approximate human ratings is that we can generate a large amount of synthetic data for these error types. In this paper, we generate $\approx 160K,$ synthetic examples for $13$ mismatch error types and use publicly available task-specific data with dataset size ranging from a few thousand to hundreds of thousands of examples with human annotation (details in Section \ref{sec:exp}).

Our approach to model human judgments involves two steps: 1) task-agnostic pre-training step where we use the synthetic examples from mismatch error types to train a shared (base) neural network model for all the 7 NLP tasks and 2) task-specific finetuning step where we finetune the pre-trained model for a specific task to predict the human ratings over different evaluation criteria. The output from the task-specific models approximates the human judgments along with the interpretable mismatches between the given pair of texts. 

Figure \ref{fig:arch} shows the architecture for the proposed mismatch-based evaluation model with pre-training and finetuning steps. 
In both these steps, we use pre-trained BERT \cite{devlin2018bert} to extract linguistic features via embeddings for both the reference and generated texts to predict the mismatch types and the human ratings (\textit{textual features}). 
We generate (2x) 64-dimensional textual features, one for the machine text and the other for the reference text. 
It is common to generate millions of synthetic examples for pre-training the neural network \cite{sellam2020bleurt} or to use tens of thousands of human-annotated data \cite{rei2020comet} to make the model robust to unseen texts. On the other hand, automatic evaluation metrics utilize handcrafted logic to compute the score on any pair of texts. In Section \ref{sec:exp}, we show that evaluation metrics as features can reliably predict these mismatch error types (\textit{scalar features}).  We demonstrate that even with a few (synthetic and task-specific) samples, our approach benefits from the handcrafted logic in the evaluation metrics to boost the prediction performance. We choose the evaluation metrics from different NLP tasks to represent different properties of natural language text. 

Unlike the textual features, the features from the automatic evaluation metrics are required to be invariant to different permutations. Traditional neural network-based models (including BERT) are very sensitive to the permutations of the input sequence. We use SetTransformer \cite{lee2019set} to extract permutation-invariant scalar features from the automatic evaluation metrics so that the scalar feature does not change under any permutation of the evaluation metric scores. We scale the evaluation metric scores between 0 and 1 before passing them to SetTransformer. We believe that textual features are extremely useful for the prediction when the reference and/or machine-generated texts are similar to the texts seen during pre-training or finetuning steps whereas scalar features are good for unseen texts. Based on this intuition, we combine the reference and generated texts with the scores computed from the automatic evaluation metrics for prediction. Both the textual and scalar features are concatenated and projected (via linear layer) to either $13$ mismatch error types for the pre-training step or human ratings on the task-specific evaluation criteria during the finetuning step.

\section{Experimental Results}
\label{sec:exp}

In this section, we show experimental results validating the proposed model for predicting human judgments.  
\begin{figure*}[ht]
    \centering
    \includegraphics[width=1\linewidth]{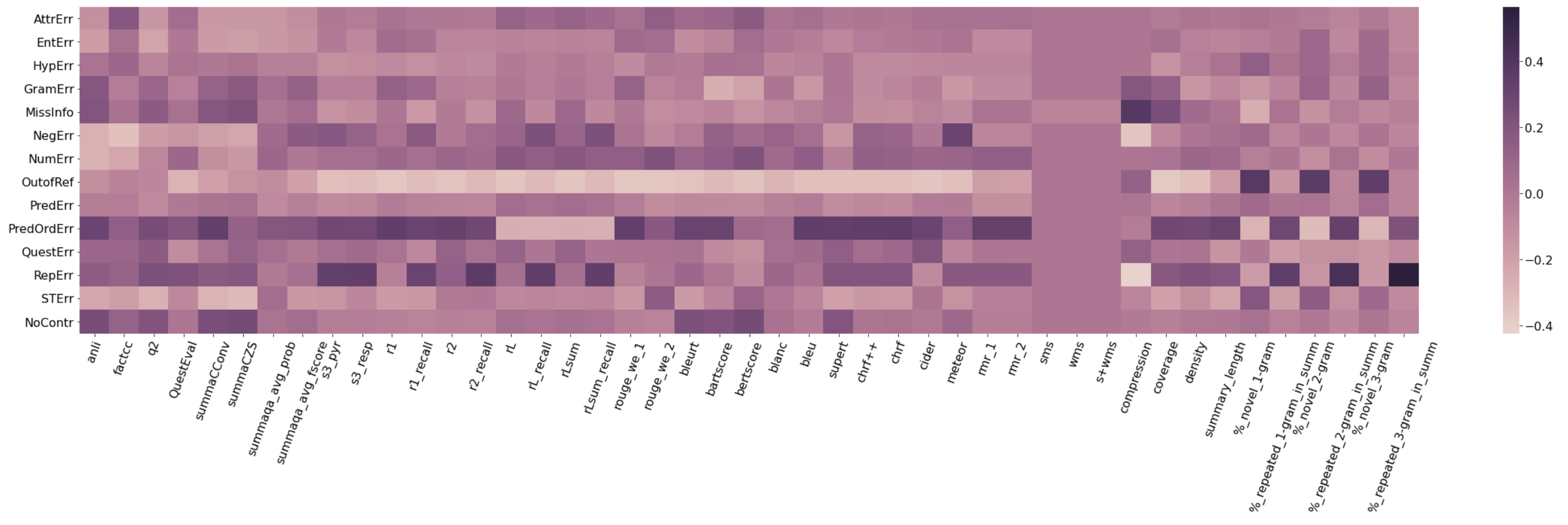}
    \includegraphics[width=0.9\linewidth]{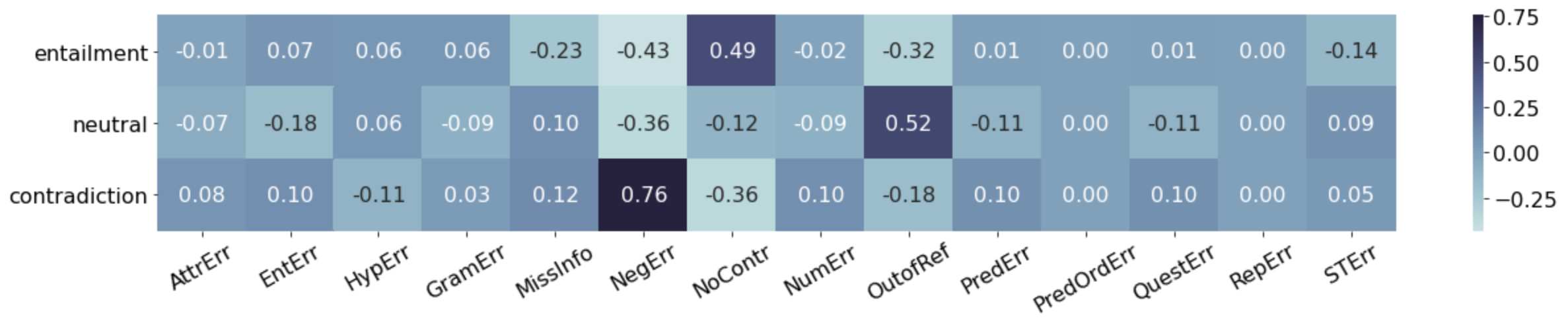}
    \caption{(Top) Correlation between mismatch error types vs automatic evaluation metrics from different NLP tasks. (Bottom) Correlation between mismatch error types vs human evaluation criteria for NLI task.}
    \label{fig:corr_autoeval}
\end{figure*}
\subsection{Datasets}
To train our proposed model based on mismatch error types to predict human judgments, we use synthetic examples for 13 mismatch error types during pre-training and real annotated examples from 7 NLP tasks for task-specific finetuning. We generate the synthetic examples by sampling the reference text from multiple NLP tasks (SQuAD \cite{rajpurkar2016squad}, WebNLG \cite{gardent2017webnlg}, MSCOCO \cite{lin2014microsoft}).  
Following the previous works \cite{sai2021perturbation,glockner2018breaking}, we use template-based perturbations on reference text to generate the synthetic examples for each mismatch type. E.g., perturbation rules to introduce subject-verb disagreement or dropping stopwords for GramErr, changing names/gender or changing the object order for EntErr, etc. In addition to the 13 error types, we include an additional \textit{No Contradiction} type (NoContr) for machine-generated text that matches the reference text. We believe this additional category helps with the better prediction of mismatch types during pre-training. We generate $\approx 200K$ synthetic examples in total for the pre-training task ($160K$ for training and the rest for validation). We report results on datasets from 7 NLP Tasks with human annotations: AS \cite{fabbri2021summeval},  IC \cite{aditya2015images}, QG \cite{nema2018towards}, MT \cite{bojar2017results}, DG \cite{mehri2020usr}, D2T \cite{gardent2017webnlg} and NLI \cite{williams2018broad} for task-specific finetuning step. We show the number of human-annotated examples used for task-specific finetuning for all 7 NLP tasks in Table \ref{tab:main_results}.

\subsection{Correlation with Mismatch Error Types}
Since most automatic evaluation metrics correlate poorly with human evaluation criteria, we study how well the proposed mismatch error types correlate with the human evaluation criteria and automatic evaluation metrics. 
Figure \ref{fig:corr_autoeval} shows the correlation plots for the proposed mismatch error types (with NoContr type). The correlations between mismatch types vs automatic evaluation metrics reveal key insights to justify the use of evaluation metrics as scalar features in our model. For instance, OutofRef is negatively correlated but PredOrdErr is positively correlated with most metrics, ngram-based metrics are highly correlated with RepErr, etc. The hardcoded logic-based evaluation metrics are equally correlated with our mismatch types as the neural network-based evaluation metrics. We also show the correlations between the mismatch types and human evaluation criteria for NLI (entailment, neutral, and contradiction). It is interesting to see that NoContr is positively correlated with entailment, NegErr is positively correlated with contradiction. These correlations will help guide the model for better prediction of human judgments on these human evaluation criteria. We also include the correlation plots for other NLP tasks in the supplementary material.

\begin{table*}[ht]
\centering
\resizebox{\textwidth}{!}{%
\begin{tabular}{|l|c|c|c|c|c|c|c|}
\hline
\textit{\textbf{Tasks}} &
  \multicolumn{1}{l|}{\textbf{AS}} &
  \multicolumn{1}{l|}{\textbf{IC}} &
  \multicolumn{1}{l|}{\textbf{QG}} &
  \multicolumn{1}{l|}{\textbf{MT}} &
  \multicolumn{1}{l|}{\textbf{DG}} &
  \multicolumn{1}{l|}{\textbf{DT}} &
  \multicolumn{1}{l|}{\textbf{NLI}} \\ \hline
\textit{\begin{tabular}[c]{@{}l@{}}\# Samples\\ (Task-Specific)\end{tabular}} &
  \begin{tabular}[c]{@{}c@{}}1600\\ (SummEval)\end{tabular} &
  \begin{tabular}[c]{@{}c@{}}2007\\ (Flickr30k)\end{tabular} &
  \begin{tabular}[c]{@{}c@{}}2726\\ (AQG)\end{tabular} &
  \begin{tabular}[c]{@{}c@{}}240,287 \\ (WMT2017-19)\end{tabular} &
  \begin{tabular}[c]{@{}c@{}}{420} \\ {(PersonaChat)}\end{tabular} &
  \begin{tabular}[c]{@{}c@{}}5,918 \\ (WebNLG)\end{tabular} &
  \begin{tabular}[c]{@{}c@{}}9,818 \\ (MNLI)\end{tabular} \\ \hline
\textit{RMSE} &
  0.18 (0.00) &
  0.23 (0.00) &
  0.16 (0.00) &
  0.19 (0.00) &
  0.26 (0.00) &
  0.24 (0.00) &
  *
   \\ \hline
\textit{Kendall's $\tau$} &
  0.30 (0.01) &
  0.49 (0.00) &
  0.52 (0.01) &
  0.35 (0.00) &
  0.31 (0.02) &
  0.39 (0.01) &
  0.89 (0.00)
   \\ \hline
\textit{Spearman's $\rho$} &
  0.41 (0.01) &
  0.62 (0.00) &
  0.66 (0.01) &
  0.50 (0.00) &
  0.40 (0.03) &
  0.49 (0.01) &
  0.92 (0.00)
   \\ \hline
\end{tabular}%
}
\caption{Model performance (agreement with human ratings) measured using Root Mean Squared Error (RMSE), Kendall's $\tau$ correlation and Spearman's $\rho$ correlation on $7$ NLP tasks (averaged over human evaluation criteria). Top row shows the dataset (in parentheses) and number of samples used for each task during finetuning step.  $^{*}$ indicates RMSE is not available as the human ratings are defined on three classes: Entailment, Neutral, Contradiction.}
\label{tab:main_results}
\end{table*}

\begin{figure*}[h]
    \centering
    \includegraphics[width=0.64\linewidth]{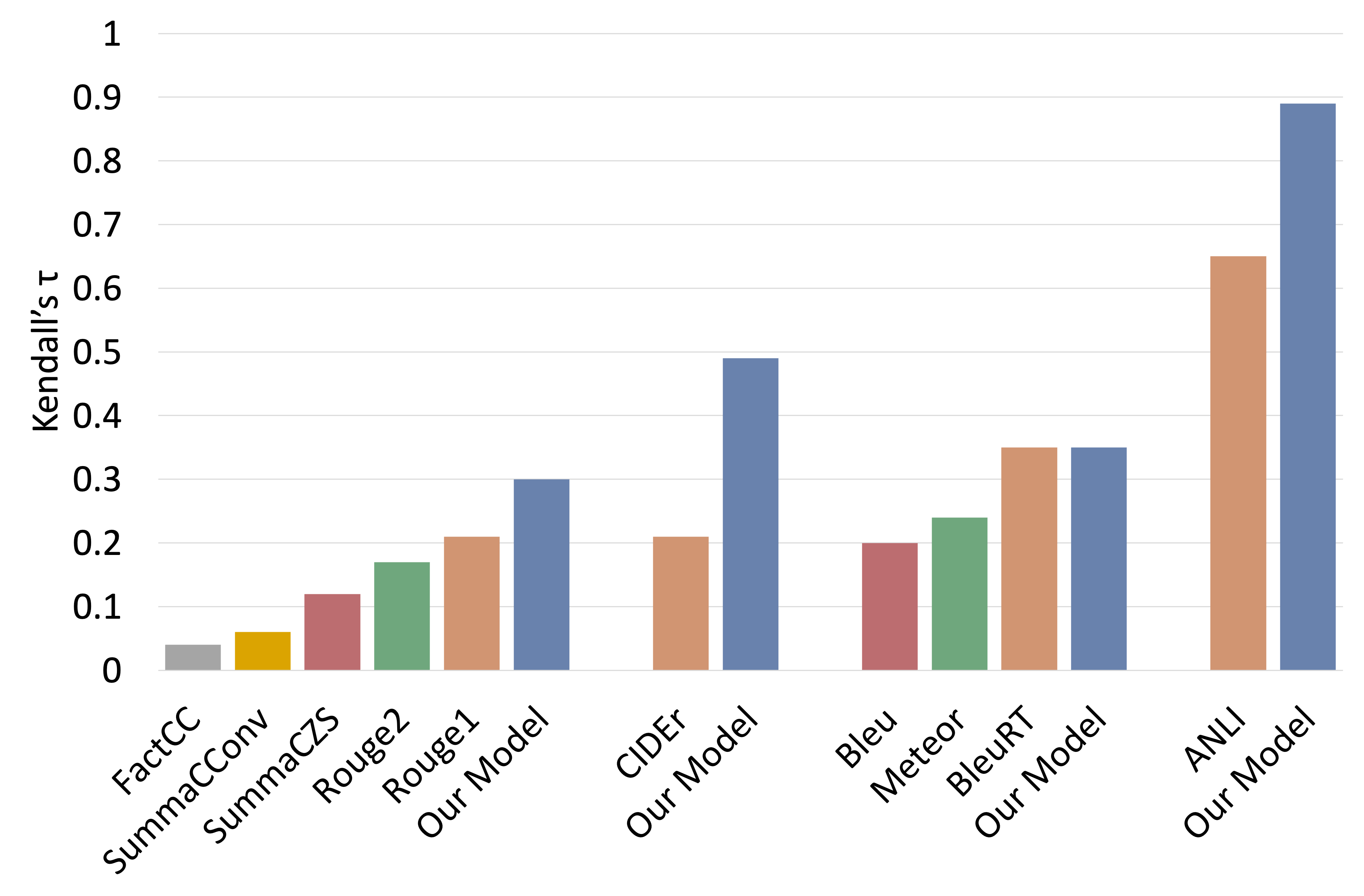}
    \caption{Comparison of the task-specific evaluation metrics  with the proposed model. Kendall's $\tau$ correlation (agreement with human ratings) is used for the performance comparison. Average over 3 runs with 100 samples randomly taken from the task-specific test set.}
    \label{fig:eval_comp}
\end{figure*}

\subsection{Model Performance}

In this section, we evaluate our model both at the task-agnostic pre-training and task-specific finetuning steps. We use an 80/20 split for both steps. We precompute the automatic evaluation metrics for the pairs of texts in both synthetic and finetuning datasets for faster computation. We use the accuracy to evaluate the performance of the pre-trained model on predicting the mismatch types; RMSE (lower is better), Kendall's $\tau$ correlation (higher is better), and Spearman's $\rho$ correlation (higher is better) between the human ratings and the predicted ratings to evaluate the performance of the task-specific finetuned models. Since we have multiple human evaluation criteria per task (e.g., entailment, neutral, and contradiction in NLI), we report the results by averaging the performance of the finetuned model over the human evaluation criteria from that task. All the experimental results reported in this paper are averaged over 3 random runs.

Table \ref{tab:main_results} shows the task-specific finetuned model performance on predicting the human rating using both the mismatch error types and scalar features.  
Our pre-trained model predicts the mismatch types on the held-out synthetic data with $98\%$ accuracy. 
We finetune the trained model on the task-specific data and achieve relatively lower RMSE scores on most of the tasks. Kendall's $\tau$ and Spearman's $\rho$ measure the linear correlation between the human ratings and the model-predicted ratings. We see that in all the NLP tasks, our model predictions align well ($\approx 0.50$ in correlation) with the human ratings. Since the NLI task involves classification labels (-1 for contradiction, 0 for neutral, and 1 for entailment) instead of human rating scores, we didn't report the RMSE score. We see that the correlation (both $\tau$ and $\rho$) for NLI is high compared to the other tasks. We believe that the NLI task is relatively easier for our proposed model compared to the other task.

Figure \ref{fig:eval_comp} compares the proposed model based on the mismatch error types against the task-specific automatic evaluation metrics both hardcoded logic-based and neural network based rules. Kendall's $\tau$ correlation between the metrics and the human ratings is used for the performance comparison. We can see that the proposed model outperforms the other metrics significantly in AS, IC and NLI. In addition, we outperform a popular neural network-based evaluation model for machine translation, BLEURT on different language pairs from both WMT2018 and WMT2019 (See supplementary for more details).

\begin{figure*}[ht]
    \centering
    \includegraphics[width=0.75\linewidth]{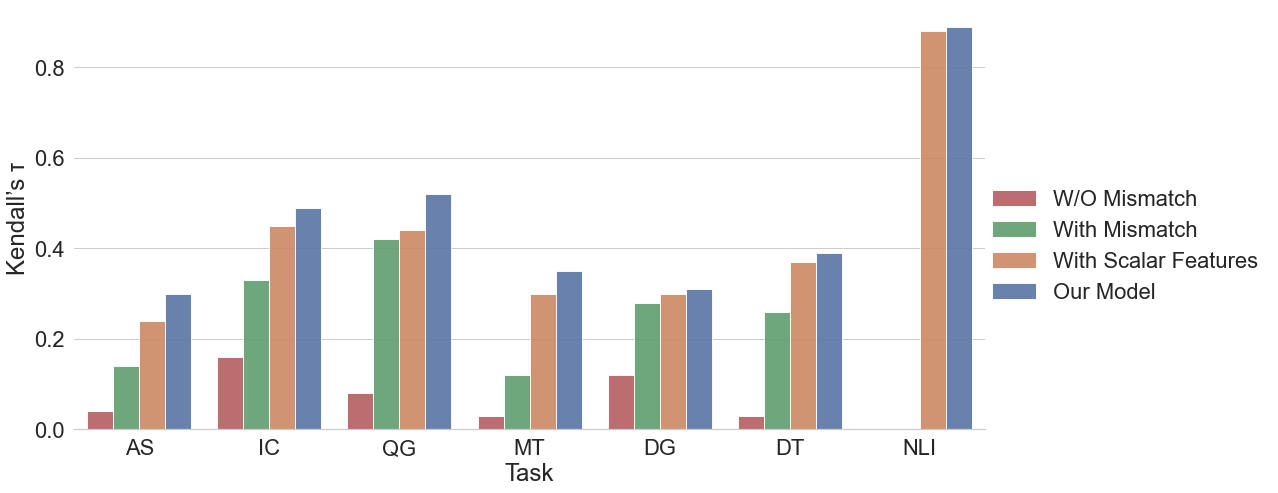}
    \includegraphics[width=0.75\linewidth]{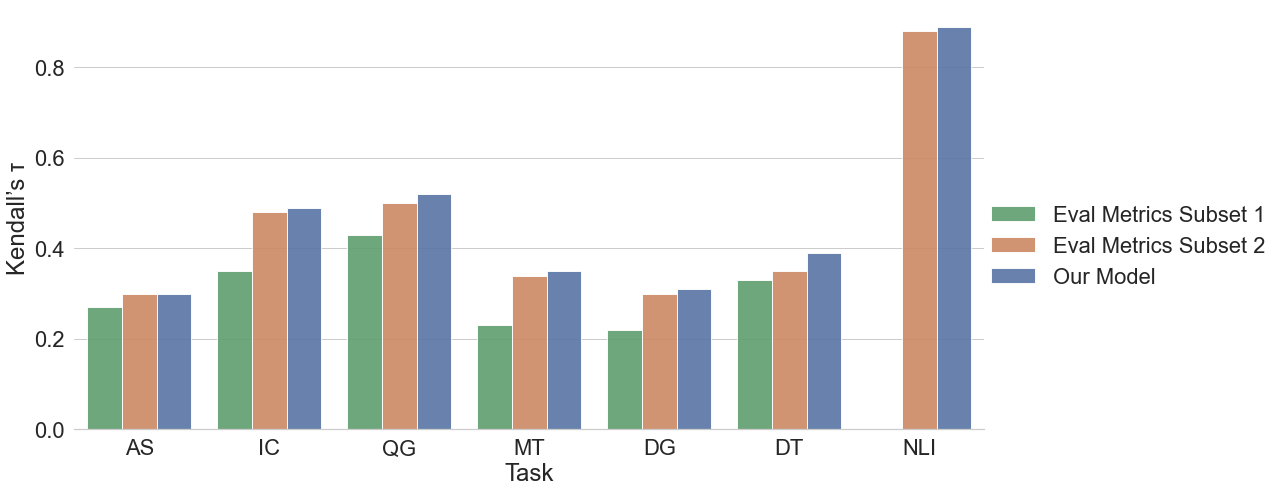}
    \caption{(Top) Agreement on human ratings using Kendall's $\tau$ correlation for different settings of the proposed approach on 7 NLP tasks. We start with Textual Features +\textit{Without mismatch} (no pre-training step with mismatch error types) and \textit{Without Scalar Features} (no evaluation metric scores as features during pre-training and finetuning steps) (Bottom) Agreement on human ratings using Kendall's $\tau$ correlation with different subset of automatic evaluation metrics used for scalar features. The two subsets are selected based on the overall cost (time and space complexity) to compute the metric scores (See table \ref{tab:featcost} in the supplementary material).}
    \label{fig:feat_comb}
\end{figure*}

\subsection{Ablation Studies}
\label{sec:ablation}
In this section, we analyze the importance of the evaluation metrics and the mismatch types for predicting task-specific human judgments. First, we compare the proposed model architecture with different settings such as with and without the mismatch error types for pre-training steps and with and without the scalar features extracted from the automatic evaluation scores using SetTransformer. Figure \ref{fig:feat_comb} (top) shows Kendall's $\tau$ correlation for the different experimental setups. 
We start with the base model that uses BERT to extract the textual features and predict the human ratings without the pre-training step for predicting mismatch error types and without the scalar features from evaluation metric scores. We write this setup as \textit{Textual Features + Without Mismatch}. Next, the baseline considers the pre-training step with mismatch types but without any scalar features.  We write this setup as \textit{Textual Features + With Mismatch}. We consider an additional baseline that uses the scalar features but without the pre-training step for mismatch-type prediction. We call this baseline, \textit{Textual Features+ With Scalar Features}. Finally, we have the proposed model that considers both the prediction step for mismatch error types and scalar features extracted from automatic evaluation metric scores.

We see that in text-only features, the pre-training step with mismatch error type significantly boosts the performance of the task-specific finetuning step, specifically in IC, QG, DG, and DT. In NLI, text-only features didn't perform as well as expected (0.0 Kendall's $\tau$ correlation). We observe that using automatic evaluation metric for scalar features significantly boost the performance of the overall model. We believe that evaluation metrics provide valuable properties (both via the hardcoded logic-based metrics such as Rouge, METEOR, etc, and neural network-based evaluation models such as ANLI, FactCC, etc) of the input texts for better performance of the fine-tuned models.  The proposed model with both the scalar features and the mismatch error types for pre-training outperforms all the other model setups. Our proposed model gets a little boost from the pre-training with mismatch error types along with the scalar features.

Figure \ref{fig:feat_comb} (Bottom) compares the importance of evaluation metric scores as a feature for the model prediction. We know from our previous experiment, evaluation metric score as scalar features provide a significant boost to our proposed model. One of the key issues with using automatic evaluation metrics as features is the cost associated with computing the scores, both space and time complexity. Time complexity measures how long it takes to compute the score for a given pair of text and space complexity measures the storage space occupied by the neural network-based evaluation model. We show the time and space complexity of each metric used in this paper in the supplementary material. To address this concern, we study the importance of cost in our model prediction. 

We choose 2 subsets of evaluation metrics with low and high costs. The subset with low-cost metrics includes hardcoded logic-based metrics such as ROUGE, METEOR, etc. The subset with high-cost metrics is mostly neural network-based models such as ANLI, FactCC, SummaC, etc.  We observe that metrics with low cost perform comparably to the metrics with high-cost as scalar features. In some tasks such as IC, QG, MT, and DT, the difference is noticeable. In NLI, the difference is significantly higher compared to any other tasks. This reveals that a subset of evaluation metrics can be selected based on the computational constraints to tradeoff between the cost and the model performances.

\begin{figure}[h]
    \centering
    \includegraphics[width=1.0\linewidth]{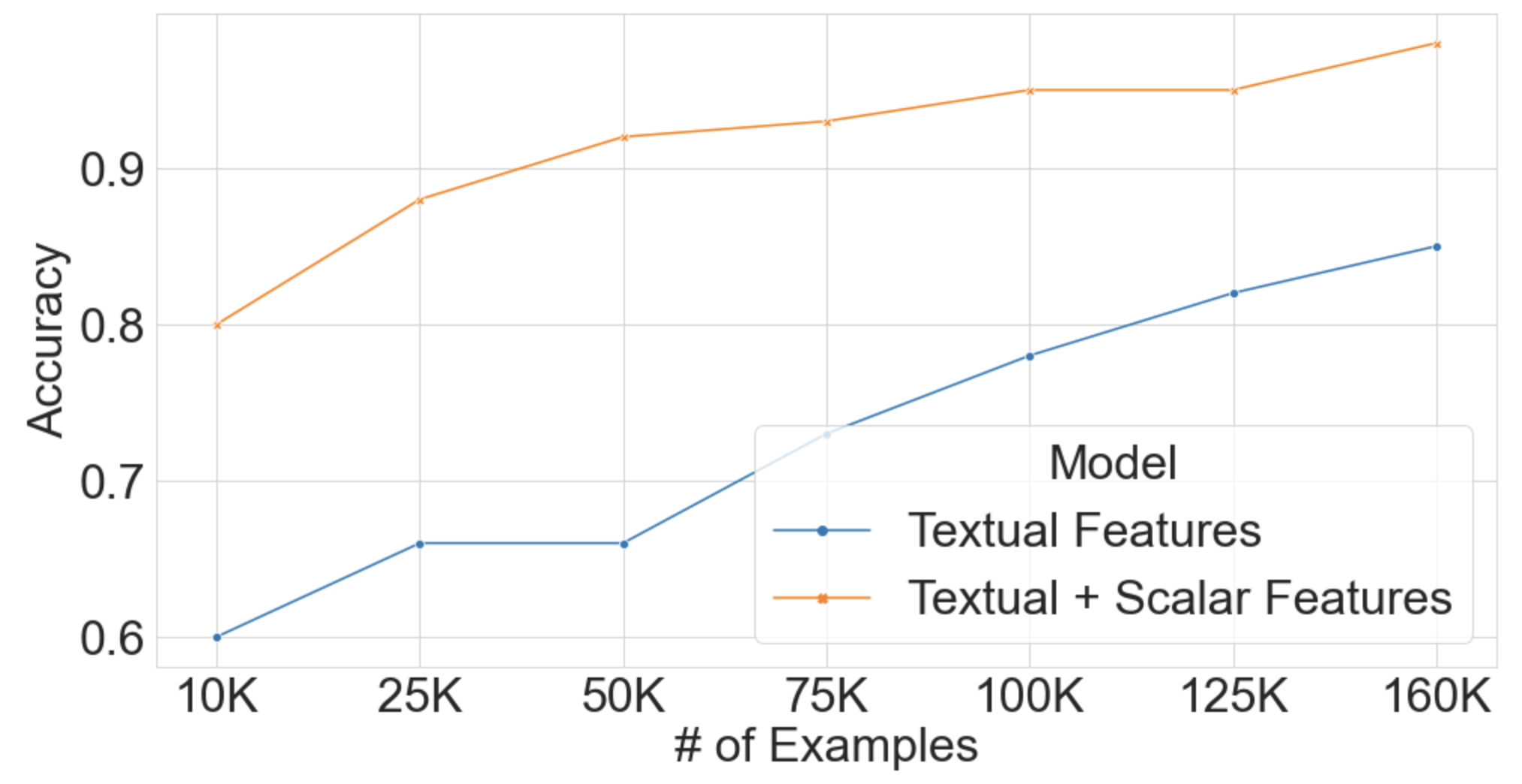}
    \caption{Performance of the pre-trained model with and w/o scalar features on different synthetic sample size. Accuracy of predicting the mismatch error types is used for comparison.}
    \label{fig:sampleeff}
\end{figure}

\begin{figure*}[ht]
    \centering
    \includegraphics[width=0.85\linewidth]{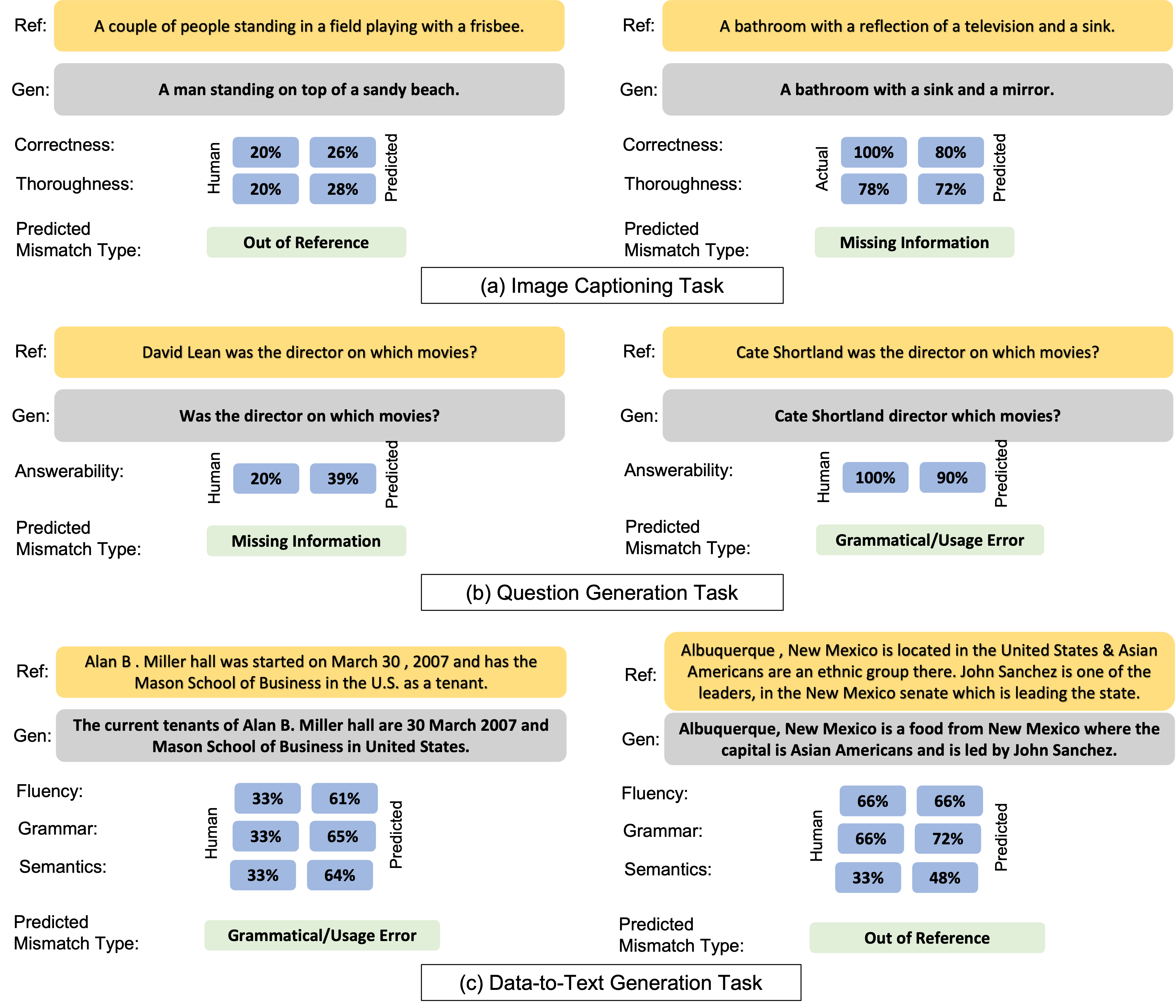}
    \caption{Sample examples taken from 3 tasks (IC, QG and DT) with both predicted mismatch error type and predicted human evaluation scores. Both the human-annotated (human) and our model-estimated (predicted) evaluation criteria scores are reported for comparison.}
    \label{fig:examples}
\end{figure*}

In Figure \ref{fig:sampleeff},  we study the importance of evaluation metrics as scalar features on sample complexity during the pre-training step. We choose different sample sizes from synthetic data for predicting the mismatch error type ranging from $10K$ to $160K$. We see the model with both the textual and scalar feature achieves better performances with a limited number of samples to train the model. This shows that evaluation metrics as scalar features have likely improved the sample complexity of the proposed model. Finally, in Figure \ref{fig:examples}, we show some sample text from 3 tasks (IC, QG and DT) showing both the predicted mismatch error type and predicted human evaluation criteria scores.

\section{Related Work}
\label{sec:related}

\textbf{Automatic evaluation metrics} such as ROUGE \cite{lin2004rouge}, BLEU \cite{papineni2002bleu}, METEOR \cite{banerjee2005meteor}, have been proposed for different tasks as a substitute for human annotations. In NLP, text generation tasks have extensively used these metrics to measure the quality of the machine-generated text. 
Evaluation metrics are either task-specific (ANLI \cite{williams2022anlizing}, SummaC \cite{laban2022summac}, CIDER \cite{vedantam2015cider}, SUPERT \cite{gao2020supert}) or task-agnostic (BERTScore \cite{zhang2019bertscore}, BLEURT \cite{sellam2020bleurt}), and are based on either human handcrafted logic (ROUGE, BLEU, METEOR) or neural framework (BERTScore, BLEURT). For human annotation, several dimensions such as coherence, consistency, and fluency are considered to measure the quality of the generated text, yet most \autoevals\ compute a single score to summarize the evaluation.
Further, these single-scored metrics often do not correlate well with the human ratings. To address this problem, several attempts have been proposed to combine multiple evaluation metrics. ROSE \cite{conroy2008mind} uses a linear combination of ROUGE variations (ROUGE\_1, ROUGE\_2, ROUGE\_L, ROUGE\_Lsum) for the machine translation task and the combined score is better than the individual rouge scores in the evaluation. $S^3$ \cite{peyrard2017learning} uses the combination of the ROUGE scores and Jenson-Shannon Divergence to predict the human rating. Neural-based approaches such as BLEURT and COMET directly train on the overall human rating for the sample texts. Even though neural-based evaluation metrics seem promising, they often require tens of thousands of training samples to mimic human rating, struggle with new domains/tasks and unseen samples, and still output a single score.

\textbf{Understanding Evaluation Metrics}: Recently, there has been growing interest in understanding what these \autoevals\ measure in terms of fine-grained evaluation criteria. 
It is done by studying different error categories (mismatches) in the machine-generated texts but claim none of the existing evaluation metrics can predict all of the mismatches, without providing any solution.
Perturbation checklist \cite{sai2021perturbation} uses template-based perturbation on multiple tasks based on the human evaluation criteria to study mismatches. FRANK \cite{pagnoni2021understanding} studies different evaluation metrics on factuality in abstractive summarization using error types. Scarecrow \cite{dou2022gpt} explores errors in prompt-based text generation by large language models. BreakingNLI \cite{glockner2018breaking} evaluates different metrics on synthetic data created from the external knowledge graph WordNet \cite{miller1995wordnet}.  \citet{tang2021confit} studies different types of factuality and hallucinations in the generated text by large language models.

In this work, we unify these task-specific research directions. We propose several evaluation models that combine the best of handcrafted logic on robust evaluation, a neural framework for text representations, and mismatch error types types to measure the quality of the generated text based on the human evaluation criteria.

\section{Conclusion}
In this paper, we proposed a neural framework for evaluating the quality of the machine-generated text w.r.t the reference text. To achieve this, we defined a set of mismatch error types to approximate the human ratings over a set of evaluation criteria. We showed that in addition to the BERT-based text representation, feature-invariant representations learned from the automatic evaluation metrics improve the prediction of both the mismatch types as well as human ratings with pre-training on only a limited amount of synthetic examples with mismatch error types. We further showed that mismatches between pairs of texts provide an interpretable way to explain human judgments, through a series of ablation studies and correlation analyses. Our proposed mismatch error types is a crucial bridge between automatic evaluation metrics and human evaluation criteria, leading to more interpretable predictions for NLP models. 

\section*{Acknowledgement}
We would like to thank Ramon Fernandez Astudillo and Tahira Naseem for their invaluable comments on an earlier version of this work.

\section*{Limitations}
One limitation of our work, which is also an avenue for future work, is that it is not fully understood yet why the mismatch error types help much more in some tasks than others. Trying to develop a more task or even instance-specific understanding of the benefits of mismatch error types will be very useful. We also want to try our proposed approach on a wider set of tasks, using different foundational models, and under the distribution shift setting to see if the mismatch error types as auxiliary supervision can improve robustness of natural language processing systems.

\section*{Ethics Statement}
With the ubiquity of natural language processing systems in real-world applications, especially in sensitive domains, it is very important that the machine-generated text is of high quality, as measured by a list of human evaluation criteria such as coherence, consistency, among others. Thus, from a societal perspective, our proposed mismatched error types provides a way to evaluate the quality of machine-generated text with respect to the reference text. From an ecological perspective, our proposed model design only involves synthetic data for pre-training and minimal computation overhead. In addition, from a trustworthiness perspective, \textsc{mismatch} provides an interpretable scheme to identify the differences between pairs of text which makes it very suitable for sensitive applications in NLP.
\newpage 
\bibliography{references}
\bibliographystyle{acl_natbib}

\newpage
\appendix

\onecolumn

\section{Mismatch Error Types Correlation Plots}

\begin{table*}[ht]
\resizebox{\textwidth}{!}{%
\begin{tabular}{p{4cm}p{8cm}p{8.2cm}p{5cm}}
\hline
\textbf{Error Type} &
  \textbf{Definition} &
  \textbf{Example Sentence} &
  \textbf{Human Evaluation Criteria} \\ \hline
\textit{Grammatical/Usage Error} &
  Faulty or incorrect use of the grammar and syntax. &
  \begin{tabular}[c]{@{}l@{}}ref:  Two paintings are on the wall.\\ gen: Two painting is on the wall.\end{tabular} &
  Fluency \\ \hline
\textit{Predicate Error} &
  \begin{tabular}[c]{@{}l@{}}Error in the predicate or its usage \\ with respect to the reference text.\end{tabular} &
  \begin{tabular}[c]{@{}l@{}}ref:  John entered the kitchen.\\ gen: John found the kitchen.\end{tabular} &
  \begin{tabular}[c]{@{}l@{}}Answerability, Relevance, \\ Making sense\end{tabular} \\ \hline
\textit{Entity Error} &
  Mismatch in the primary arguments of the predicate. &
  \begin{tabular}[c]{@{}l@{}}ref:  A dog chased a cat.\\ gen: A dog chased a rat.\end{tabular} &
  \begin{tabular}[c]{@{}l@{}}Relevance, Correctness, \\Thoroughness, Informativeness,\\ Referential clarity\end{tabular} \\ \hline
\textit{Predicate Ordering Error} &
  \begin{tabular}[c]{@{}l@{}}Error in causal or temporal ordering of \\ the predicates/events.\end{tabular} &
  \begin{tabular}[c]{@{}l@{}}ref: The police arrested the suspect \\ then he was taken to prison.\\ gen: The suspect was taken to the prison\\ then the police arrested him.\end{tabular} &
  \begin{tabular}[c]{@{}l@{}}Flow/Coherence, Answerability, \\ Making sense, repetitions\end{tabular} \\ \hline
\textit{Hyponyms/ Hypernyms Errors} &
  Violations in hypernym/hyponym usage. &
  \begin{tabular}[c]{@{}l@{}}ref: Jim studied mechanical engineering.\\ gen: Jim studied architectural science.\end{tabular} &
  Informativeness \\ \hline
\textit{Numerical Error} &
  \begin{tabular}[c]{@{}l@{}}Error in numerical, quantifiers or related to numbers \\ (ordinals, cardinals, etc)\end{tabular} &
  \begin{tabular}[c]{@{}l@{}}ref:  Martha ate four apples.\\ gen: Martha ate six apples.\end{tabular} &
  Correctness, Thoroughness \\ \hline
\textit{Spatial/Temporal Error} &
  \begin{tabular}[c]{@{}l@{}}Error in spatial or geographic information \\ (location, time, etc).\end{tabular} &
  \begin{tabular}[c]{@{}l@{}}ref:  Dave lives in south Chicago.\\ gen: Dave lives in south Chile.\end{tabular} &
  Correctness, Thoroughness \\ \hline
\textit{Attribute/Modifier Error} &
  \begin{tabular}[c]{@{}l@{}}Mistakes in additional information \\ concerning the predicates and entities.\\ (not covered by numerical, spatial, geographic)\end{tabular} &
  \begin{tabular}[c]{@{}l@{}}ref:  Greg has two small dogs.\\ gen: Greg has two big dogs.\end{tabular} &
  Correctness, Thoroughness \\ \hline
\textit{Question Error} &
  Error/change in the nature of the question's intention. &
  \begin{tabular}[c]{@{}l@{}}ref:  Did you take the dog to the vet?\\ gen: When did you take the dog to the vet?\end{tabular} &
  Answerability \\ \hline
\textit{Negation} &
  Negated compared to the reference text. &
  \begin{tabular}[c]{@{}l@{}}ref: Susan took the gift. \\ gen: Susan did not take the gift.\end{tabular} &
  Adequacy \\ \hline
\textit{Missing Information} &
  Missing key details from the reference text. &
  \begin{tabular}[c]{@{}l@{}}ref: Bob drove to the hospital and saw a doctor.\\ gen: Bob saw a doctor.\end{tabular} &
  Adequacy, Thoroughness, Data Coverage \\ \hline
\textit{Out of Reference} &
  Contains additional details not present in the reference text. &
  \begin{tabular}[c]{@{}l@{}}ref: Jack and Jane are friends.\\ gen: Jack and Jane are friends. Jack plays football.\end{tabular} &
  Adequacy, Making Sense, Listening \\ \hline
\textit{Redundant/Repetition} &
  Same/similar information repeated more than once. &
  \begin{tabular}[c]{@{}l@{}}ref:  Tom met Sam at the party.\\ gen: Tom went to the party. Tom met Sam at the party.\end{tabular} &
  \begin{tabular}[c]{@{}l@{}}Thoroughness, Avoid Repetition, \\ Data coverage\end{tabular} \\ \hline
\end{tabular}
}
\caption{Mismatch Error types between the reference and model-generated texts.}
\label{tab:errortypes}
\end{table*}

\begin{figure*}[h!]
    \centering
    \includegraphics[width=1\linewidth]{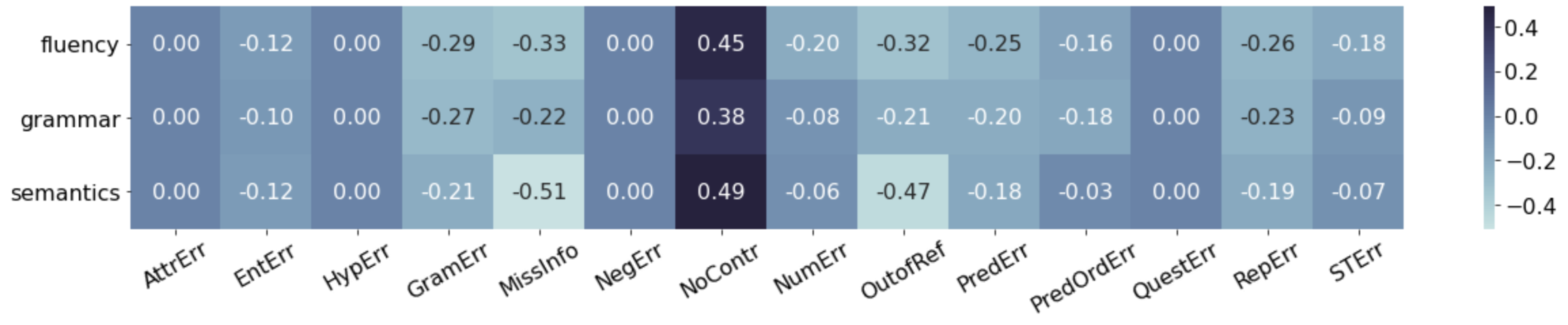}
    \includegraphics[width=1\linewidth]{imgs/correlations/nli.png}
    \includegraphics[width=1\linewidth]{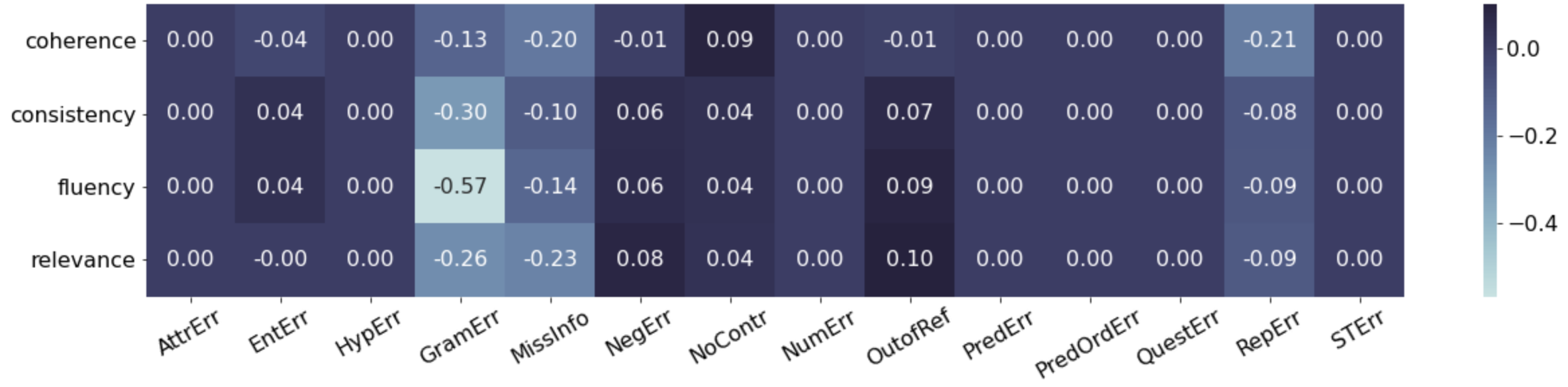}
    \includegraphics[width=1\linewidth]{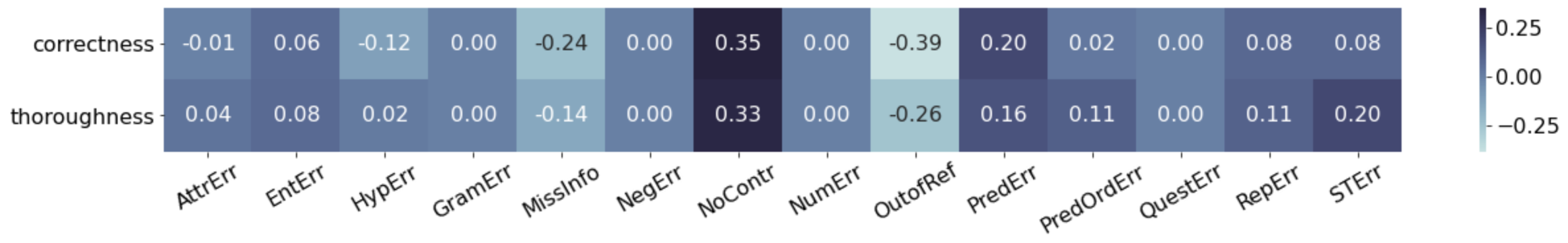}
    \includegraphics[width=1\linewidth]{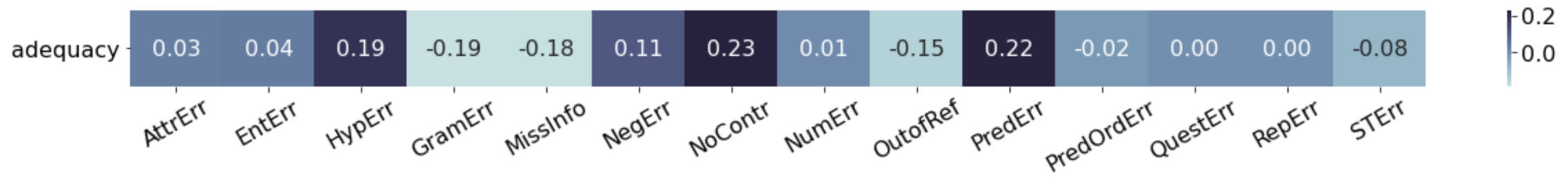}
    \includegraphics[width=1\linewidth]{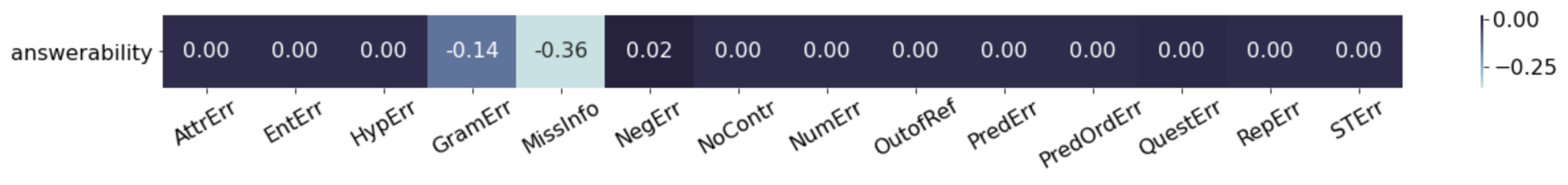}
    \caption{Correlation between mismatch error types vs human evaluation criteria for 7 NLP tasks: Data-To-Text, Natural Language Inference, Abstractive Summarization, Image Captioning, Machine Translation and Question Generation. }
    \label{fig:corr_humaneval}
\end{figure*}

In Table \ref{tab:errortypes}, we present our proposed mismatched error types, their definitions, and examples, and the corresponding human evaluation criteria they aim to capture. In Figure \ref{fig:corr_humaneval}, we present the correlation plots between the mismatch error types and the human evaluation criteria for 7 popular NLP tasks. We see a significant correlation between several of the mismatch error types with the human evaluation criteria, especially the ones they aim to capture, across the different tasks.

\section{Comparison with BLEURT on WMT Shared Metric Task}
In this section, we compared the proposed neural evaluation framework against BLEURT, a popular neural network-based evaluation metric in Machine Translation. BLEURT is a strong baseline for our proposed approach where they used automatic evaluation metrics such as ROUGE, BLEU, and BERTScore as pre-training signals for the auxiliary task. Unlike our proposed approach for 7 NLP tasks, the BLEURT evaluation metric is primarily used for evaluating generated texts by the machine translation models. BLEURT uses $1.8$ million synthetic examples from Wikipedia for pretraining whereas our proposed approach uses $\approx 160K$ synthetic examples from datasets such as SQUAD, WebNLG, MSCOCO, etc. The BLEURT metric relies on the linguistic (text) features extracted from the reference and machine-generated texts, whereas, our proposed approach uses both the text features and the scalar features extracted from the automatic evaluation score. 

In Tables \ref{tab:bleurt_results_2018} and \ref{tab:bleurt_results_2019}, we compare the BLEURT results with our mismatch-based evaluation approach on the 2018 and 2019 WMT Metric Shared Task. We see that on both datasets, we outperform BLEURT on all language pairs in terms of Kendall's Tau correlation. This shows that fine-grained evaluation criteria based on mismatch error types are better auxiliary signals than automatic evaluation metrics.

\begin{table*}[h]
\centering
\begin{tabular}{lcccccccc|}
\\ \hline
\multicolumn{1}{|l|}{\textbf{\begin{tabular}[c]{@{}l@{}}Models/\\ Languages\end{tabular}}} &
  \multicolumn{1}{c|}{\textbf{cs-en}} &
  \multicolumn{1}{c|}{\textbf{de-en}} &
  \multicolumn{1}{c|}{\textbf{et-en}} &
  \multicolumn{1}{c|}{\textbf{fi-en}} &
  \multicolumn{1}{c|}{\textbf{ru-en}} &
  \multicolumn{1}{c|}{\textbf{tr-en}} &
  \multicolumn{1}{c|}{\textbf{zh-en}} &
  \multicolumn{1}{c|}{\textbf{avg}}
  \\ \hline
\multicolumn{1}{|l|}{\textit{BLEURT}}       & \multicolumn{1}{c|}{35.6}   & \multicolumn{1}{c|}{44.2} &  \multicolumn{1}{c|}{40.0} &  \multicolumn{1}{c|}{32.1} &  \multicolumn{1}{c|}{31.9} &  \multicolumn{1}{c|}{35.5} &  \multicolumn{1}{c|}{29.7} &  \multicolumn{1}{c|}{35.6}\\ \hline 

\multicolumn{1}{|l|}{\textit{Our Model}}       & \multicolumn{1}{c|}{36.0}   & \multicolumn{1}{c|}{44.7} &  \multicolumn{1}{c|}{40.6} &  \multicolumn{1}{c|}{33.3} &  \multicolumn{1}{c|}{32.5} &  \multicolumn{1}{c|}{36.0} &  \multicolumn{1}{c|}{30.4} &  \multicolumn{1}{c|}{36.9}\\ \hline 
\end{tabular}
\caption{Agreement with human ratings on the WMT18 Metrics Shared Task. Kendall Tau ($\tau$) is used to evaluate results.}
\label{tab:bleurt_results_2018}
\end{table*}

\begin{table*}[h]
\centering
\begin{tabular}{lcccccccc|}
\\ \hline
\multicolumn{1}{|l|}{\textbf{\begin{tabular}[c]{@{}l@{}}Models/\\ Languages\end{tabular}}} &
  \multicolumn{1}{c|}{\textbf{de-en}} &
  \multicolumn{1}{c|}{\textbf{fi-en}} &
  \multicolumn{1}{c|}{\textbf{gu-en}} &
  \multicolumn{1}{c|}{\textbf{kk-en}} &
  \multicolumn{1}{c|}{\textbf{lt-en}} &
  \multicolumn{1}{c|}{\textbf{ru-en}} &
  \multicolumn{1}{c|}{\textbf{zh-en}} &
  \multicolumn{1}{c|}{\textbf{avg}}
  \\ \hline
\multicolumn{1}{|l|}{\textit{BLEURT}}       & \multicolumn{1}{c|}{31.2}   & \multicolumn{1}{c|}{31.7} &  \multicolumn{1}{c|}{28.3} &  \multicolumn{1}{c|}{39.5} &  \multicolumn{1}{c|}{35.2} &  \multicolumn{1}{c|}{28.3} &  \multicolumn{1}{c|}{42.7} &  \multicolumn{1}{c|}{33.8}\\ \hline 

\multicolumn{1}{|l|}{\textit{Our Model}}       & \multicolumn{1}{c|}{31.6}   & \multicolumn{1}{c|}{32.3} &  \multicolumn{1}{c|}{28.1} &  \multicolumn{1}{c|}{40.5} &  \multicolumn{1}{c|}{35.4} &  \multicolumn{1}{c|}{28.3} &  \multicolumn{1}{c|}{45.8} &  \multicolumn{1}{c|}{33.9}\\ \hline 
\end{tabular}
\caption{Agreement with human ratings on the WMT19 Metrics Shared Task. Kendall Tau ($\tau$) is used to evaluate results.}
\label{tab:bleurt_results_2019}
\end{table*}

\section{Sample Examples with Intrepretable Mismatch Error}
In Tables \ref{tab:ic_examples}, \ref{tab:qg_examples} and \ref{tab:nli_examples}, we show sample examples from the task-specific dataset. One of the advantages of our proposed methods is that in addition to predicting the human rating based on the different human evaluation criteria, it can provide the mismatches occurred between the reference and machine-generated text for further interprtation of the predicted human ratings.

\section{Additional Details}
In Table \ref{tab:eval_mat}, we show the list of all the automatic evaluation metrics used in our proposed model along with the associated NLP task with their references.
Table \ref{tab:featcost} shows the cost associated with computing the evaluation metric scores. It includes both the time complexity (in seconds) and space complexity (in MBs). Time complexity measures how long will the metric takes to compute the evaluation score, whereas space complexity measures how much storage space this metric will consume during the training process. We can see that the hardcoded logic-based metrics such as ROUGE, METEOR, etc are relatively low-cost compared to the neural network-based models such as ANLI, FactCC, etc with the high cost.

\begin{table*}[h!]
\resizebox{\textwidth}{!}{%
\begin{tabular}{p{8cm}p{8cm}p{4cm}p{4cm}p{4cm}p{4cm}p{5cm}}
\hline
\textbf{Reference Text} &
  \textbf{Model Generated Text} &
  \textbf{Correctness} &
  \textbf{Thoroughness} &
  \textbf{Predicted Correctness} &
  \textbf{Predicted Thoroughness} &
  \textbf{Predicted Mismatch}\\ \hline
   a mounted police officer riding down a city street past parked cars & a man riding a horse on a city street & 1 & 1 & 0.82 & 0.74 & Error in Hyponyms or Hypernyms\\ \hline
  a couple of people standing in a field playing with a frisbee. &
  a man standing on top of a sandy beach & 0.2 & 0.2 & 0.26 & 0.28 & Out of Reference \\ \hline
  a bathroom with a reflection of a television and a sink. & a bathroom with a sink and a mirror & 1 & 0.8 & 0.78 & 0.72 & Missing Information \\ \hline
  \end{tabular}
}
\caption{Example sentences from Image Captioning task with predicted human evaluation criteria and mismatch type.}
\label{tab:ic_examples}
\end{table*}

\begin{table*}[h!]
\resizebox{\textwidth}{!}{%
\begin{tabular}{p{8cm}p{8cm}p{4cm}p{4cm}p{5cm}}
\hline
\textbf{Reference Text} &
  \textbf{Model Generated Text} &
  \textbf{Answerability} &
  \textbf{Predicted Answerability} &
  \textbf{Predicted Mismatch}\\ \hline
   How do co-teachers work with each other to fulfill the needs of students? & When do co-teachers work with each other to fulfill the needs of students? & 1 & 0.89 & Question Error\\ \hline
  David Lean was the director on which movies ? &
  was the director on which movies ? & 
  0.2 & 0.39 & Missing Information \\ \hline
  Cate Shortland was the director on which movies ? & Cate Shortland director which movies ? & 1 & 0.90 & Grammatical \& Usage \\ \hline
  \end{tabular}
}
\caption{Example sentences from Question Generation task with predicted human evaluation and mismatch type.}
\label{tab:qg_examples}
\end{table*}

\begin{table*}[h!]
\resizebox{\textwidth}{!}{%
\begin{tabular}{p{8cm}p{8cm}p{4cm}p{4cm}p{5cm}}
\hline
\textbf{Reference Text} &
  \textbf{Model Generated Text} &
  \textbf{Label} &
  \textbf{Prediction} &
  \textbf{Predicted Mismatch}\\ \hline We got him out of that 10-20 range. & Got him just out of 9,000-9,500 now.
   & Contradiction &
  Contradiction & Numerical or Quantifiers \\ \hline
  How about when you were in school? &
  How about when you went to bed? & 
  Contradiction & Contradiction & Entity Error \\ \hline
  \end{tabular}
}
\caption{Example sentences from NLI task with predicted human evaluation and mismatch type.}
\label{tab:nli_examples}
\end{table*}

\begin{table*}[h!]
\resizebox{\textwidth}{!}{%
\begin{tabular}{p{8cm}p{8cm}p{4cm}p{4cm}p{4cm}p{4cm}p{4cm}p{4cm}p{5cm}}
\hline
\textbf{Reference Text} &
  \textbf{Model Generated Text} &
  \textbf{Fluency} &
  \textbf{Grammar} &
  \textbf{Semantics} &
  \textbf{Predicted Fluency} &
  \textbf{Predicted Grammar} &
  \textbf{Predicted Semantics} &
  \textbf{Predicted Mismatch}\\ \hline
   alan b . miller hall was started on march 30 , 2007 and has the mason school of business in the u . s . as a tenant &
  the current tenants of alan b. miller hall are 30 march 2007 and mason school of business in united states & 0.33 & 0.33 & 0.33 & 0.61 & 0.65 & 0.64 & Grammatical \& Usage\\ \hline
  albuquerque , new mexico is located in the united states and asian americans are an ethnic group there . john sanchez , is one of the leaders , in the new mexico senate which is leading the state &
  albuquerque, new mexico is a food from new mexico where the capital is asian americans and is led by john sanchez. & 0.66 & 0.66 & 0.33 & 0.66 & 0.72 & 0.48 & Out of Reference \\ \hline
  \end{tabular}
}
\caption{Example sentences from Data-to-Text task with predicted human evaluation criteria and mismatch type.}
\label{tab:d2t_examples}
\end{table*}

\begin{table*}[ht]
\centering
\begin{tabular}{|l|l|l|l|}
\hline
\textbf{Metric} & \textbf{Task} &  \textbf{Reference} \\ \hline
ANLI & Natural Language Inference  & \cite{nie2020adversarial} \\ \hline
factcc & Abstractive Summarization & \cite{kryscinski2020evaluating} \\ \hline
Q2 & Knowledge-grounded Dialogue &  \cite{honovich2021q2} \\ \hline
QuestEval & Abstractive Summarization &  \cite{scialom2021questeval} \\ \hline
summaC (Conv/CZS) & Abstractive Summarization   & \cite{laban2022summac}  \\ \hline
summaqa (avg\_prob/avg\_fscore) & Abstractive Summarization  & \cite{scialom2019answers} \\ \hline
$S^3$ (pyr/resp) & Abstractive Summarization  & \cite{fabbri2021summeval} \\ \hline
ROUGE & Summarization   & \cite{lin2004rouge} \\ \hline
ROUGE-WE & Abstractive Summarization & \cite{ng2015better} \\ \hline
BLEURT & Machine Translation  & \cite{sellam2020bleurt} \\ \hline
BARTScore & Text Generation  & \cite{yuan2021bartscore} \\ \hline
Blanc & Machine Translation  & \cite{lita2005blanc} \\ \hline
Bleu & Machine Translation  & \cite{papineni2002bleu} \\ \hline
SUPERT & Multi-document summarization  & \cite{gao2020supert} \\ \hline
chrf & Machine Translation & \cite{popovic2015chrf} \\ \hline
chrf++ & Machine Translation & \cite{popovic2017chrf++} \\ \hline
Cider & Image Description Evaluation & \cite{vedantam2015cider} \\ \hline
Mauve & Text Generation & \cite{pillutla2021mauve} \\ \hline
METEOR & Machine Translation & \cite{banerjee2005meteor} \\ \hline
RMR (1/2) & Abstractive Summarization & \cite{zhu2021enhancing} \\ \hline
sms/wms/s+wms & Distance between documents & \cite{kusner2015word} \\ \hline
coverage/density & Text summarization & \cite{grusky2018newsroom} \\ \hline

\end{tabular}
\caption{Complete list of automatic evaluation metrics used in this paper.}
\label{tab:eval_mat}
\end{table*}

\begin{table*}[ht]
\centering
\begin{tabular}{|l|l|l|}
\hline
\multicolumn{1}{|c|}{\textbf{Eval Metrics}} &
  \multicolumn{1}{c|}{\textbf{\begin{tabular}[c]{@{}c@{}}Time\\Complexity\\(\textit{sec})\end{tabular}}} &
  \multicolumn{1}{c|}{\textbf{\begin{tabular}[c]{@{}c@{}}Space\\Complexity\\(\textit{MegaBytes})\end{tabular}}} \\ \hline
\textit{ANLI}         & 426   & 2163.88  \\ \hline
\textit{BARTScore}    & 94    & 1883.47  \\ \hline
\textit{BERTScore}    & 66    & 2107.17  \\ \hline
\textit{Blanc}        & 7048  & 3081.04  \\ \hline
\textit{BLEU}         & 47    & 169.12   \\ \hline
\textit{BLEURT}       & 869   & 1949.05  \\ \hline
\textit{CHRF}         & 80    & 161.43   \\ \hline
\textit{CIDER}        & 14    & 238.25   \\ \hline
\textit{\begin{tabular}[c]{@{}l@{}}Datastats\\ (n-gram, etc)\end{tabular}} &
  81 &
  399.64 \\ \hline
\textit{FactCC}       & 165   & 2197.16  \\ \hline
\textit{MAUVE}        & 16191 & 3380.99  \\ \hline
\textit{METEOR}       & 65    & 249.09   \\ \hline
\textit{Q2}           & 18790 & 6883.66  \\ \hline
\textit{QuestEval}    & 7996  & 5493.33  \\ \hline
\textit{RMR}          & 2012  & 167.24   \\ \hline
\textit{ROUGE}        & 1542  & 181.41   \\ \hline
\textit{ROUGE\_we1}   & 32830 & 1219.65  \\ \hline
\textit{ROUGE\_we2}   & 34218 & 1207.79  \\ \hline
\textit{$S^3$  (pyr/resp)}           & 458   & 10470.42 \\ \hline
\textit{SMS}          & 20972 & 289.70   \\ \hline
\textit{SummaC\_conv} & 287   & 2323.56  \\ \hline
\textit{SummaC\_zs}   & 284   & 2125.90  \\ \hline
\textit{SummaQA}      & 923   & 4830.03  \\ \hline
\textit{SUPERT}       & 532   & 2237.44  \\ \hline
\end{tabular}
\caption{The feature costs in terms of the time taken and memory associated with each evaluation metric. Time complexity includes both the data preparation and computation time (in seconds). Space complexity includes average memory taken by the evaluation metrics (in Mb).}
\label{tab:featcost}
\end{table*}

\end{document}